**Title:**

Artificial Intelligence in PET: an Industry Perspective

**Authors:**

1. Arkadiusz Sitek, Ph.D., Sano Centre for Computational Medicine, Cracow, Poland, Nawojki 11 Street, 30-072 Kraków, Poland, a.sitek@sanoscience.org
2. Sangtae Ahn, Ph.D. GE Research
3. Evren Asma, Ph.D., Canon Medical Research
4. Adam Chandler, Ph.D. Global Scientific Collaborations Group, United Imaging Healthcare, America
5. Alvin Ihsani, Ph.D., NVIDIA
6. Sven Prevrhal, Ph.D., Philips Research Europe,
7. Arman Rahmim, Ph.D., Departments of Radiology and Physics, University of British Columbia, Provincial Medical Imaging Physicist, BC Cancer, Canada
8. Babak Saboury, MD, MPH, DABR, DABNM, Department of Radiology and Imaging Sciences, Clinical Center, National Institutes of Health, Department of Computer Science and Electrical Engineering, University of Maryland Baltimore County, Baltimore, MD, USA, Department of Radiology, Hospital of the University of Pennsylvania,
9. Kris Thielemans, Ph.D., Institute of Nuclear Medicine, UCL, London, UK, Algorithms and Software Consulting Ltd, London, UK

**Corresponding Author:**

Arkadiusz Sitek

**Key Points:**
- Industry faces unique challenges in order to bring AI to PET clinical workflows

- There new AI ecosystems created to facilitate use of AI in clinics
- New computing ecosystems can include reconstructions of vendor neutral format raw PET list-mode data
- Custom workflows including image reconstructions and list-mode data processing can be used in new AI ecosystems

**Synopsis:**

A*rtificial intelligence* (AI) has significant potential to positively impact and advance medical imaging, including positron emission tomography (PET) imaging applications. This paper provides an overview of these industry-specific challenges for the development, standardization, commercialization, and clinical adoption of AI in PET, and explores the potential enhancements to PET imaging brought on by AI in the near future. In particular, the combination of on-demand image reconstruction, AI, and custom designed data processing workflows may open new possibilities for innovation which would positively impact the industry and ultimately patients.

**Abbreviation list:**

AI - artificial intelligence

LM - list mode

TOF - time of flight

PPI - Parent PET image

PI - PET image

PPI-AI - Parent PET image artificial intelligence

SLM - standardized list mode

ECG - electrocardiogram

AMC - academic medical center

PHI - protected health information

PACS - picture archiving and communication system

SUV - standardized uptake value

VOI - volume of interest

CNN - convolutional neural network

**Glossary:**

Artificial intelligence, convolutional neural networks, Positron Emission Tomography, Radiology, AI ecosystem, AI workflows, federated learning, industry perspective, data acquisition, list-mode data, standardization, cost, data access, robustness, underspecification of AI model, clinical value, regulations, AI failures, adversarial attacks, uncertainty estimation, explainability, decision making, decision support, human-machine decision making, liability, custom data processing, adoption of AI, trust in AI recommendations, PET list-mode standardization, standardized image reconstruction.

Must read:

Nice to read:

*Abstract*

A*rtificial intelligence* (AI) has significant potential to positively impact and advance medical imaging, including positron emission tomography (PET) imaging applications. AI has the ability to enhance and optimize all aspects of the PET imaging chain from patient scheduling, patient setup, protocoling, data acquisition, detector signal processing, reconstruction, image processing and interpretation. AI poses industry-specific challenges which will need to be addressed and overcome to maximize the future potentials of AI in PET. This paper provides an overview of these industry-specific challenges for the development, standardization, commercialization, and clinical adoption of AI, and explores the potential enhancements to PET imaging brought on by AI in the near future. In particular, the combination of on-demand image reconstruction, AI, and custom designed data processing workflows may open new possibilities for innovation which would positively impact the industry and ultimately patients.

# 1 Introduction

The recent popularity of *artificial intelligence* (AI) heralded as a game changing technology has generated high hopes for breakthrough advancements and changes across the entire healthcare industry. The specific area of clinical positron emission tomography (PET) imaging is no exception. In this work, we provide an industry perspective on specific opportunities and challenges for PET arising by the emergence of AI and deep learning methods.

Deep learning (DL) [1] is a machine learning technique which uses deep neural networks to create a variety of models which can process raw data. In recent years, DL has demonstrated significantly promising results for several PET applications, including segmentation, reconstruction, outcome modeling, decision support, etc. [2-8].

In this work, a *manufacturer* is defined as an industry member manufacturing PET scanners, and a *vendor* as an industry member providing AI and other processing software solutions. These two groups are not exclusive. In the present work, we sometimes interchangeably use the terms *deep learning* and *artificial intelligence*, although AI (and its subset machine learning) are wider fields. The recent spike of interest in AI is due to increased popularity of DL, especially the use of convolutional neural networks, which is why we use this convention in this paper.

The paper is organized as follows. Section 2 identifies selected challenges in the adoption of AI from the industry point of view. They are general and not PET specific. The goal of this section is not to discuss potential solutions to those challenges but rather paint a perspective on specific challenges from the industry point of view. In section 3, specific applications of AI in PET are discussed in more detail. In particular a concept of reconstruction of the list-mode data on demand is combined with AI algorithms and presented.

# 2 Challenges for commercialization

One of the main concerns for industry is to release reliable, extensively tested and validated products that impact disease and patient management. For the purpose of this article, we define a reliable product as working as intended and within a set of predefined specifications. Equally important, the product should demonstrate clinical utility. In this section we discuss some of the major concerns and obstacles the industry has to overcome.

*2.1 Development and clinical evidence*

2.1.1 Access to data

Obtaining large amounts of data to develop AI products is challenging, and often ownership of the data is not with the industry. The need for obtaining sufficient amounts of data for training, which encompass all expected variations in the data, i.e. population-based variations (both locally and geographically), body locations, disease state variations (including normal/non-disease cases) etc., adds further challenges. *Federated learning* (FL) is an approach that may at least partially alleviate the issue. In FL, AI models are trained based on data that never leave the medical institutions [9] and therefore data security and privacy are much less of a concern. The paradigm of federated learning is being widely explored (e.g. the work on FL from the London Medical Imaging & AI Centre for Value-Based Healthcare [10,11]).

### 2.1.2 Ground truth

In some applications of AI such as supervised learning, obtaining ground truth will present a great challenge. Ground truth can be obtained by an independent measurement (e.g. biopsy, post-mortem analysis), clinical outcomes (death, morbidity), previous diagnoses (e.g. radiology reports), or new reads or annotations can be utilized. Ideally, the data sets should be large, but new reads and annotations make data preparation a lengthy and expensive process.

### 2.1.3 Robustness

Of particular commercial interest is a reliable, regulatory cleared product that performs according to specifications regardless of geographical location, patient mix and local preferences and guidelines. Unfortunately, AI algorithms can generalize poorly and are dependent on the data sets used to train and test the algorithms. An AI algorithm may produce unreliable results if characteristics of the input deviate from the training data. This has critical consequences. It is acceptable to publish an AI algorithm tested on homogeneous data (e.g. from a single or small number of institutions using well-defined study inclusion criteria) as long as those limitations are transparently disclosed in the publication. However, a commercially available product ought to be applied to real life data that may be more diverse and complex than single center study data, which may render certain limitations of an algorithm as non-acceptable. In general radiology, there are many large publicly available datasets which can be used to test generalization of developed AI algorithms. Unfortunately, there are few such sets available that include PET data, making the development of AI algorithms for PET more difficult.

### 2.1.4 Underspecification

Another obstacle to generalization of AI is a recently documented problem of *underspecification* [12]. This term denotes the problem that if we train the same model a number of times with slightly different initial weights on the same large dataset and achieve similar performance on the test set, there is no guarantee that those models will perform the same in the real world. This is a very difficult problem to tackle and extremely important from the industry point of view as the real-world performance is what matters. When many models are trained on the same set of data with random initial weights and applied to a certain unseen real-world scenario, some models may work and some other models may not. At the phase of model development, it is difficult to tell which of those models will work and which ones will not. Testing models with diverse real-world data will alleviate the problem although not entirely. Therefore, we emphasize the importance of post-market surveillance after algorithm deployment which becomes even more important compared to classical (not deep learning) offerings.

### 2.1.5 Clinical value

When commercializing AI algorithms, there is a need to demonstrate that the product provides clinical value and evidence that supports the intended use. To generate such data that can be utilized as evidence for potential regulatory claims that translate into customer value, often multi-center, multi-reader studies are required. Here, we emphasize that often one develops an excellent technological solution to a clinical problem but when introduced to clinical workflows it is not widely used in routine clinical practice by clinicians. Appropriately designed external evaluation studies at clinical sites by clinicians could mitigate the problem.

*2.2 Regulatory pathways*

AI's towering dependence on data exposes MedTech´s regulatory and privacy challenges more than ever before: compounded by the sharp teeth that GDPR has afforded the EU, with global effects, academia and industry are only now learning to safely share massive amounts of data.

Regulatory bodies, too, increasingly demand being shown the data used to train the AI parts of software submitted for their approval. However, basing approval on the data creates the conundrum as once approved, retraining with new data would invalidate it and burden industry and administration with incessant re-approval cycles. Luckily, everybody agrees a solution is direly needed. In the USA, the FDA is working on an action plan, and the EU has just released a white paper with very similar thoughts [13-15]. Obviously, an eventual world-wide joint framework will be key for industry and data-owning individuals alike.

*2.3 Return on investment*

The healthcare industry requires a reasonable return of investment in order to create or sustain a viable business. For applications of AI in PET products investment in development should be properly justified by balancing the growth potential of the AI technology with the considerable risks. AI may require a non-traditional business model in which subscription approaches, architectures open to third parties like marketplaces and new ecosystems are used (see section 3 of this article). It is an industry challenge to figure out why and how clients would pay for AI innovations.

*2.4 Understanding AI*

### 2.4.1 Explainability

Explainability is an important factor associated with the adoption of AI from the commercial point of view. In short, in deep learning methods the decisions made by AI are often opaque, black box decisions. For more details on this problem please refer to [16]. For AI algorithms to succeed in the commercial world, the users of the algorithms have to gain trust in them. For example, a clear explanation on how the AI algorithm arrived at a certain classification can increase trust in the subsequent clinical decisions which AI recommends.

2.4.2 Education and trust

It is critically important to educate users about AI's capabilities and even more importantly its limitations. Most current applications of neural networks are some form of image denoising where very noisy images, presumably from short or low dose scans, are converted into images that appear less noisy. However, this does not mean that nothing is lost due to shorter or lower dose scans. Users need to understand that the quantitative lesion/ROI performances of their images are still governed by the statistics of the acquisition. AI can mimic longer or higher dose scans by making backgrounds smoother but cannot create the information that is lost due to shorter or lower dose scans. Nonetheless, we note that AI can improve image quality such as lesion detectability or signal-to-noise ratio by using better priors, system models, data correction, or noise models learned from data.

Another related topic is how clinicians determine the reliability of lesion SUVs. They may look at how noisy a large, approximately constant region such as the liver is and decide that smoother regions indicate more reliable lesion quantitation. For typical reconstruction algorithms such as OSEM this approach works reasonably well because if the noise correlation lengths are short, the single-image-noise in the liver is related to the standard deviation of a single liver voxel. That is in turn related to the standard deviations of individual lesion voxels which finally determine the standard deviation of the lesion SUV. In contrast, when the background is smoothed using AI-based methods, this connection is lost. The single-image-noise in the liver may be greatly reduced without any significant changes to the lesion ensemble noise properties. Therefore a clinician looking at an image denoised with neural networks should be cautious about interpreting the variability (or uncertainty) of the lesion SUV. It should also be noted that denoising could introduce an additional bias in the lesion SUV.

2.4.3 Combining human and AI insights

In the foreseeable future, human decision makers will be augmented/assisted and not replaced by automated algorithms. Unavoidably there will be situations where a human opinion is different from that of an algorithm. This creates opportunities and challenges. Opportunities because the combination of AI and humans may create a better and more accurate decision [17,18]. However, it creates a problem on how to meaningfully combine human and AI insights. The final decision in the foreseeable future will be made by humans and some solutions are needed to deal with disagreements. One such approach could be that AI provides explanations or examples from the past of similar images with known outcomes, which may persuade the physician. Another resolution of such conflicts could be that we teach the AI algorithm to consider the physician's arguments for the different opinion (similarly as the difference in opinions is resolved between two physicians) and then to recompute the estimates. It is however unclear how this can be accomplished in the current workflows, and requires future research. These are important ethical issues of paramount importance to industry which need to be resolved with cooperation with stakeholders including clinical and ethical experts, patient advocacy groups, governmental bodies, and of course the industry [19]. Finally, we anticipate that, when AI makes a clinical decision without human intervention some day perhaps in the not-too-distant future, we will face a complex problem of who is liable for a wrong decision made by AI, similar to self-driving car liability.

*2.5 Failures*

2.5.1 Critical failures

If algorithms do not perform according to specifications, it constitutes a major problem for vendors. For example, DL-based image reconstruction can be unstable resulting in severe artifacts [20]. This risk is often a consequence of the poor generalization of the AI algorithms and the fact that results presented by an AI algorithm are often not explainable. If a spectacular error is made by AI, it is very damaging to the perception of a product even if it works within specified characteristics. When publishing a paper, the same penalty is applied if the algorithm had an error or a spectacular error. However, when we deploy an AI-product, a spectacular failure could be much more detrimental to the trust in the algorithm. These types of errors, although very damaging, are very hard to mitigate with the current state of knowledge about neural networks. On a positive note, as much as critical failures of AI are damaging to its reputation, they are at least easily identifiable as errors. There are some safety features that can be used ("graceful failure"). For example, if we use AI to compute quantitative values and if the computed values are outside of the physiological range, one may display a message that AI failed to compute the value rather than providing it to the user. For classification problems these types of mitigations are much more difficult to implement. This is certainly important from an industry point of view and an important direction of future research.

2.5.2 Uncertainty estimation

Clear communication to the interpreting physician of uncertainty in the AI result is crucial in building trust in the AI system, since, as previously discussed, no AI system will be perfect or able to handle the huge range of real world inputs. It is not practical, or even possible, for AI developers to aim for a perfect result every time, so communication of uncertainty is of paramount importance[21,22]. Large uncertainty alerts about low confidence in provided inference. This is particularly true for nuclear imaging techniques which produce data with high noise compared to other modalities, and this noise may translate to uncertainty in reconstructed images and AI decisions. Suppose we develop an AI algorithm which automatically detects the volume of interest (VOI) of abnormal uptake of FDG. Ideally the algorithm would also provide an estimation of uncertainty on the VOI size and SUV. This uncertainty can be expressed by providing a range of values that with a high likelihood contains the true value (confidence intervals). This can also be done using Bayesian approaches where each value of the volume or SUV is assigned probability of being true [23]. Estimation of such uncertainty can be accomplished with neural networks using approximations to Bayesian approaches [24,25] or some other approximate methods [26,27].

### 2.5.3 Malicious AI, adversarial attacks

Another potential concern is that AI and deep learning methods either by accident or maliciously may introduce perturbations in the images. Some of these perturbations can be imperceptible to humans but may have a drastic effect on AI outcomes. For example, in an image manufactured by malicious AI, the analyzing AI may detect a tumor with 100% certainty which remains completely invisible to a human observer. For more on adversarial attacks refer to [28]. When used for PET image reconstruction, AI may also introduce perturbations with image textures different from those obtained by standard iterative methods, which may be misinterpreted as abnormalities.

*2.6 To err is human. How does this apply to AI?*

Another issue that industry faces is to roll out products that will over time earn the trust of radiologists and nuclear medicine physicians and convince them to use algorithmic advice. We already drew the reader's attention to challenges associated with explainability. *Algorithm aversion* is another, potentially more serious, obstacle which may prevent seamless acceptance of AI solutions. Human decision makers are averse to algorithmic predictions after seeing them perform; even with evidence of non-inferiority of the AI algorithm, humans still tend to follow advice given by humans because people more quickly lose confidence in algorithms than in human forecasters after observing them repeating a mistake [29]. Algorithmic aversion may be a major obstacle to adoption of AI. AI algorithms used in augmenting human decision making will likely have to be held to very high standards by enforcing inter- and extra-user reproducibility. If we can, we should also provide quantitative values with confidence measures (see also section 2.2). Confidence is also important for yes/no or other classification decision tasks and some type of confidence measures should always be provided.

## 3 Looking into the future of AI in PET

Various academic medical centers (AMC) worldwide are currently investing to incorporate artificial intelligence (AI) in both research and clinical research settings as a prelude to AI-supported clinical workflows. For those applications, AI is largely used to scale and automate data analysis for large cohorts in multi-year studies whereby thousands of images are analyzed retrospectively. In clinical research, AI is typically used for clinical decision support as a "second opinion" to that of the clinician, to increase the saliency of structures and functions of interest in images while increasing efficiency of acquisition, and/or to identify possible regions or planes of interest in images so the clinician may improve diagnosis, increase efficiency and minimize fatigue [30-32].

The important question for the industry is how we bring AI into the clinical workflow in an efficient and scalable way. In section 3.1 we consider using AI during PET data acquisition. In section 3.2 we explore new AI ecosystems already proposed elsewhere [33-35] and discuss how to leverage the uniqueness of PET raw data (e.g. list mode) in such ecosystems.

*3.1 AI during PET-data-acquisition*

AI offers a whole new array of promising approaches that have the potential to optimize the utility of PET imaging by adjusting controllable parameters based on the specific patient-anatomy, patient-physiology, and scanner type. The basic idea of how to achieve this is summarized in figure 1. We present the ability of AI algorithms to combine various types of information to provide just-in-time inferences which help to create high fidelity PET data at the PET scanner while data is being acquired.

Below, we provide example scenarios of how such AI inferences can be applied.
**Scenario 1**: While the data is being acquired during a single bed position, AI analyses the *data acquired* (figure 1) and uses criteria of *acceptable data quality* to determine if a sufficient number of counts was acquired up to this moment. An example of what problem this may partially solve is patient motion. If AI detects substantial patient motion it triggers additional time for data acquisition also informing the operator. **Scenario 2:** Suppose we scan a patient to determine whether the SUV in a given VOI changed vs. the SUV measured in a previous PET scan. We provide the AI the *previous PET/CT scan*, *data acquired*, and maximum threshold for uncertainty of a decision (figure 1). We want to know if the SUV increased/decreased by 20% with 95% certainty. AI analyzes the data and computes the maximum possible certainty that can be reached and the additional acquisition time to reach it.

The PET scanner is also a location where manufacturer-specific AI can be deployed. Once the raw data is created and the image is reconstructed, an AI algorithm can generate insights which can be sent to PACS or other destinations along with the data. Such solutions may be very effective as the manufacturer controls the type of data the AI algorithm is exposed to. The downside is of course that it is limited to individual manufacturers.

*3.2 Vendor-neutral data processing (VNDP) platforms*

An effective approach to deployment of AI in radiology and other clinical environments is unclear. It is likely however that in the near future we will have hundreds of AI algorithms approved for use in clinics and operating on different parts of clinical workflow and data. If we do not have a common platform to deploy them and rather depend on each AI vendor to use their own methods, the deployment and growth of AI in PET could stall as the complexity quickly becomes unmanageable.

To address this, new vendor-neutral data processing platforms (VNDPs) are proposed [33-35]. In radiology the VNDPs are interfaced with PACS. AI and other algorithms can process the data pulled from PACS and other hospital IT systems. After processing, the output can be sent back to the PACS, be saved on different archives, or displayed as shown in figure 2. We will not discuss these workflows in detail here, and refer the reader to references available on this topic (vid. [33-35]). Software units that operate in VNDPs can be stacked together if their output/input type fits. Once stacked, blocks can be replaced by different blocks or stack of blocks. Such an architecture has a similarity with those used by smartphones [34] because software units are "sitting" on the platform and are activated if the "right" data arrives and they can be swapped/updated on user requests. Using this analogy we will refer to the software units as 'apps' (figure 2).

To provide an example of data processing in a VNDP platform let us consider figure 2 and processing by apps 4, 5, and 6. The input consists of PET/CT images. App 4 segments the liver using CT, app 5 detects liver lesions using PET and CT, and app 6 performs diagnosis and computes SUV using PET and CT if lesions were detected. Note that outputs from apps 4 and 5 are used by app 6.

3.2.1 Extended VNDP platform - processing standardized list mode (SLM)

Archiving PET data in a raw list-mode format has many advantages as it gives the ability to retrospectively reconstruct images on-demand. There are many examples of where such flexibility is beneficial. For example, when training AI algorithms, it allows the developer to create a larger variety of images in terms of resolution and noise from just a single raw datafile. It also allows the developer to vary the total number of counts simulating different doses. The LM format may contain information about deposited energy and time-of-flight per event information, exact crystal pairs in which the gamma photons were detected, which may lead to development of improved reconstruction algorithms or correction algorithms compared to histogrammed (sinogram) data. Since timing information is available for each event it allows for various patient motion corrections.

The availability of LM data in new ecosystems would open opportunities to processing PET data, training new AI algorithms, and deriving AI inferences. The data reconstruction in such an ecosystem would just be another processing app which can be inserted in the processing pipeline (recon apps in figure 4). An example of such processing could be, for example, raw LM data correction for randoms or scatter which could be done before image reconstruction. Ideally, in such an ecosystem one would like to standardize the format of LM data to make it easier for vendors to develop apps which would work directly on LM data irrespective of the type of scanner the data was generated on. We refer to such a format as standardized LM (SLM) format. To perform state-of-the-art reconstruction, reconstruction applications need information including geometry, detector calibration, sensitivity, etc which would have to be included in the SLM. We note that such a format does not exist at the time of writing this paper as each scanner vendor uses a proprietary format. We note that standards for raw data are long established in SPECT [36,37], and more recently MRI [38]. The SLM format for PET needs to be designed and approved by all stakeholders. A first step towards this goal would be that manufacturers disclose non-sensitive parts of their file formats, as some have already agreed to in the context of open source projects [39-41].

3.2.2 Extended VNDP platform - processing parent PET image (PPI)

Although SLM in the VNDP platform provides enormous flexibility in constructing custom processing pipelines, handling list mode files presents challenges. They are very large files (3-20 GB) and storage and network demands are considerable. Each vendor has proprietary highly optimized software which reconstructs images directly from the LM or sinograms created from the LM. Reconstruction software may have specific computing hardware requirements that may not be readily available in the VNDP platform.

AI offers an alternative approach towards creating a practical platform for generalization of the reconstruction process across different scanners and manufacturers without explicitly using LM files. The suggested approach gives up some generalizability compared to the SLM approach described in section 4.2.1, but it is more practical and well suited for use within a VNDP platform. We refer to this concept as the *parent PET image* (PPI) and summarize it in figure 4.

The main idea is that instead of handling SLM in the new ecosystem as shown in figure 3, we reconstruct on the scanner a parent image (or images) and use it instead of SLM in the VNDP platform. In the VNDP platform PPI is then used to generate on-demand various child images (figure 4). The generation of child images from the PPI is done using deep convolutional neural networks (CNNs) referred to in this paper as PPI-AI. The PPI-AI are types of apps in the ecosystem (figure 2) that convert PPIs to child images.

The PPI can be for example the high-fidelity image. PPI can actually also be a set of images, such as hi-fidelity images with and without attenuation correction, resolution modelling etc. If time-of-flight is available, it could also contain back-projections at different angles, as used by the DIRECT method [42]. There are many possibilities on how to define PPI and research is needed to determine which of those choices would be optimal. The PPI is reconstructed on the scanner and it is stored in PACS possibly along with some child images. The PPI can be pulled to the VNDP platform and almost instantaneously converted to any child image as the inference using the PPI-AI CNN model is fast. Once converted to a child image it can be further processed by AI apps or other apps as a regular PET image (figure 5).

Looking at figure 4, the PPI can be converted to a high-fidelity image, the best utility image that a vendor can generate from the LM file. When training AI apps to be used in new ecosystems (figure 2) we would like to use images of various quality with various artifacts for the app to be more robust and general. A PPI-AI model can be trained to generate poorer quality images from the PPI. Examples of such are shown as different noise/resolution tradeoff and lower dose child images in figure 4.

There are ongoing efforts to harmonize and standardize results obtained on different scanners [43,44]. This can also be done using the PPI by creating harmonized child images. For this, we would require collaboration between vendors to create a single PPI-AI CNN model which could generate harmonized images from PPIs of different vendors. We can take this concept further and imagine a situation where the user points a cursor on a lesion when viewing a high-fidelity image and transparently to the user the standardized image is created in the background and the viewing system displays standardized SUV values.

The fifth child image example provided in figure 4 is physiological motion (e.g. respiratory) correction using PPI-AI. If no motion correction is applied, regions of the PPI with motion will appear blurry. PPI-AI models can be trained to recover resolution from blurred PPIs.

Alternatively, a PPI could contain several images, e.g., in different motion states, or one in end-expiration and one without motion correction, from which a fully motion corrected image can be produced.

The training of PPI-AI models is conceptually straightforward. Suppose we want to create a PPI-AI model that generates from the PPI a half-dose image. First, we identify a training set which contains, say 1000 PET scans from some patient population. Next, we reconstruct those 1000 images from LM data using only half of the counts available in the LM. Then, we create PPIs by reconstructing images using all counts and high-fidelity reconstructions. We train neural networks (PPI-AI) with PPIs as the input and half-dose images as the target. This completes the creation of the PPI-AI model. Similarly, any other PPI-AI model can be trained. In the above we assumed that high-fidelity reconstruction image is the PPI, but this may not necessarily be the optimal choice as already discussed.

A disadvantage of using the PPI compared to SLM is that the PPI contains less information than the LM file. Timing information is not available and although the PPI can in general be a dynamic (or ECG-gated) sequence it cannot be time reframed to a different sequence. We also do not have access to deposited energy, time of flight etc. However, we remember that some of the information is transferred to PPI-AI models during training. Intuitively, during PPI-AI inference when child images are generated from the PPI, not only the information in the PPI is used but also the information "stored" in PPI-AI models.

Another disadvantage of PPI is that if a manufacturer improves the tomographic reconstruction algorithm and wants to update it on the scanner, all PPI-AI models have to be retrained which could be an automated process but it is computationally intensive. If novel reconstruction is to be applied retrospectively to data acquired in the past the PPIs have to be updated as well.

## 4 Summary

In section 2 we presented important challenges for creating and adopting AI solutions in clinics from the point of view of the industry. In section 3 we concentrated on PET explored unique to PET applications of AI during data acquisition. We examined a flexible and scalable ecosystem for deployment of AI and described a synergy of such systems with an idea of standardized list mode data and the other solution presented here based on the parent PET image concept.

There are emerging new workflows and data ecosystems in radiology. In addition to facilitating AI deployment they provide a tremendous opportunity for the PET community to transform the current paradigm of PET data processing.

## 5 Acknowledgements

Authors would like to thank Sven Zuehlsdorff, Ph.D., Siemens Medical Solutions USA, Inc., Hoffman Estates, IL, USA, sven.zuehlsdorff@siemens-healthineers.com for his input. The opinions expressed by authors in this paper may not necessarily represent the official opinions of their employers. This publication is partly supported by the European Union's Horizon 2020 research and innovation programme under grant agreement Sano No 857533 and the International Research Agendas programme of the Foundation for Polish Science, co-financed by the European Union under the European Regional Development Fund.

**BOX points:**

- Reconstruct PET data on-demand (e.g. just before or during reading) from raw data i.e. standardized list-mode data or parent PET images
- Use raw data as a part of the "patient medical record"
- Include "DICOM push" for ease of raw data transfer/storage/management
- Archive raw data which is essential for future improved recon, such as motion correction or harmonization/comparison with prior
- Process raw data in new AI ecosystems

**Figures:**

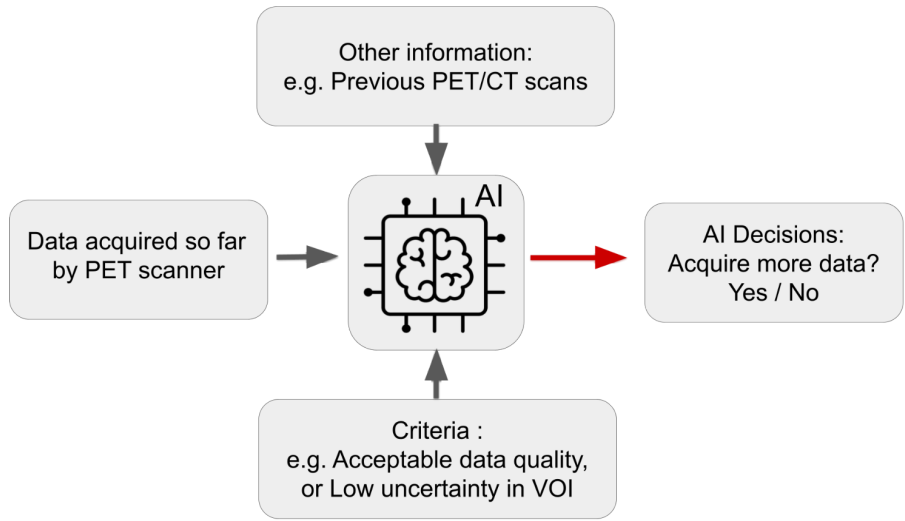

Figure 1. Conceptual depiction of AI used at the PET scanner during data acquisition. Gray arrows indicate input to AI (data acquired so far, other data acquired in the past, and criteria for decision making) and red indicate output from AI.

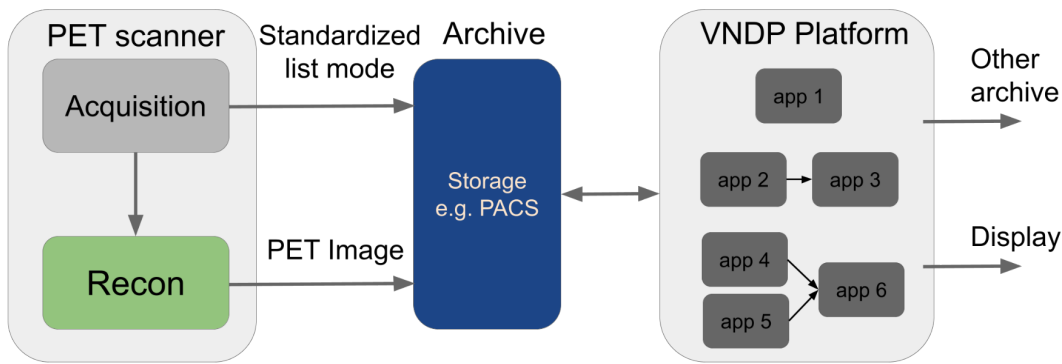

Figure 2. Simplified PET data flows in new AI ecosystems. Data stored in e.g. PACS is pulled to the VNDP platform where they are processed by software units ('apps'). Apps can be used as a single processor (app 1) or stacked (e.g. app 2 and 3 or apps 4, 5, and 6). VNDP platforms allow for creation of custom workflows with custom apps. Output from VNDP platform can either be sent back to original storage, other archive, or displayed. Interactions with hospital information systems and other sources of information are omitted for clarity. Applications of AI before data reaches storage are not shown.

Sitek *et al*, PET Clinics (2021), accepted                                                                              20

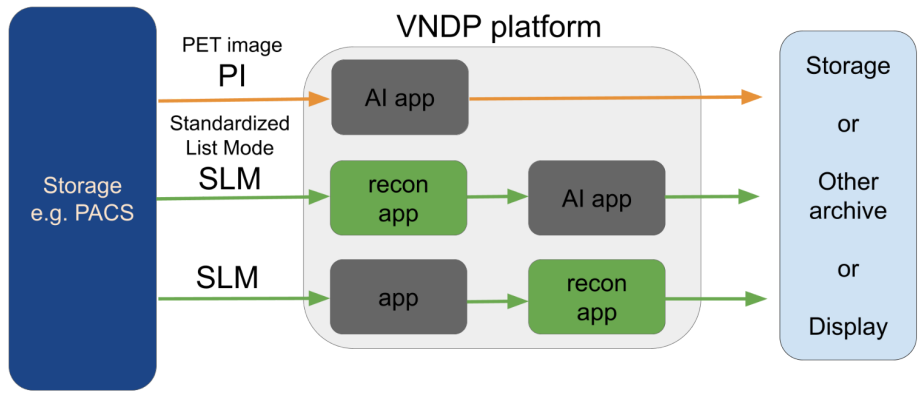

Figure 3. PET Data flow in VNDP. Orange arrows show dataflow in the new AI ecosystem with standard PET images processed by a single AI app. Green arrows show new PET-specific dataflows proposed in this work. (3) SLM can be reconstructed by recon app and processed by AI app. In the third example, SLM is pre-processed (e.g. randoms correction) and then reconstructed by the recon app. Interactions with hospital information systems and other sources of information are omitted for clarity.

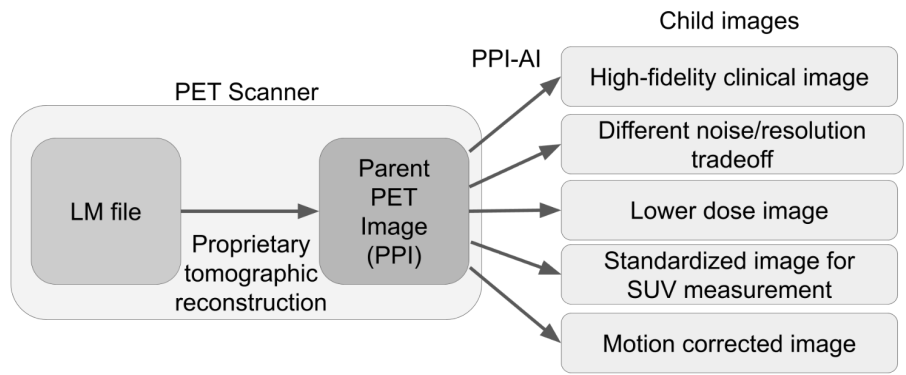

Figure 4. LM data is reconstructed using vendor-specific proprietary software at the scanner. Each manufacturer creates manufacturer-specific AI models (PPI-AI) to transform the parent PET image to child images needed for various clinical and research tasks.

Sitek *et al*, PET Clinics (2021), accepted 21

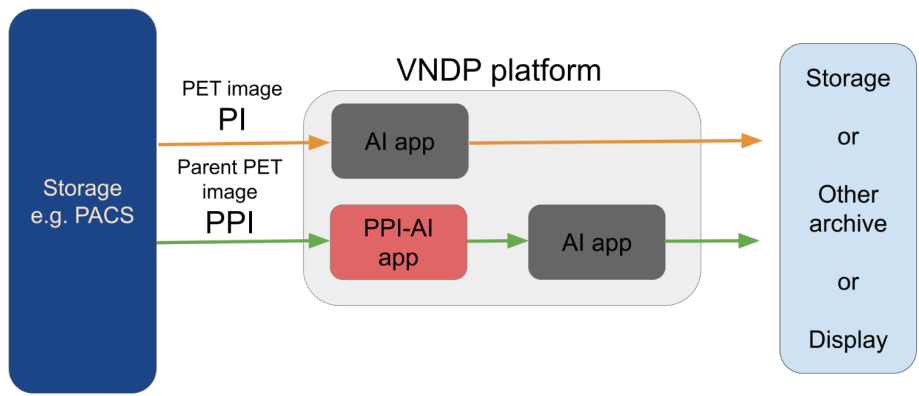

Figure 5. PET Data flow in VNDP with parent PET image (PPI) concept. Orange arrows show dataflow in the new AI ecosystem with standard PET images processed by a single AI app. Green arrows show new PET-specific dataflows proposed in this work. PET parent image is pulled from PACS and converted by PPI-AI app to a PET image which is processed by a single AI app. Interactions with hospital information systems and other sources of information are omitted for clarity.